\documentclass{article}
\usepackage{times}
\usepackage{soul}
\usepackage{url}
\usepackage[hidelinks]{hyperref}
\usepackage[utf8]{inputenc}
\usepackage[small]{caption}
\usepackage{graphicx}
\usepackage{amsmath}
\usepackage{amsthm}
\usepackage{booktabs}
\usepackage{ijcai20}
\usepackage[ruled,vlined]{algorithm2e}
\usepackage{multirow}
\usepackage{subcaption}
\usepackage{natbib}
\usepackage{hyperref}
\usepackage{url}
\usepackage{graphicx,wrapfig}  
\usepackage{nccmath}
\usepackage[dvipsnames]{xcolor}
\usepackage{amsmath}

\newcommand{\MN}{VNEstruct\xspace}

\usepackage{amsmath,amsfonts,bm}

\def\Figref#1{Figure~\ref{#1}}

\def\eqref#1{equation~\ref{#1}}
\def\Eqref#1{Equation~\ref{#1}}

\def\1{\bm{1}}

\DeclareMathAlphabet{\mathsfit}{\encodingdefault}{\sfdefault}{m}{sl}
\SetMathAlphabet{\mathsfit}{bold}{\encodingdefault}{\sfdefault}{bx}{n}

\DeclareMathOperator*{\argmin}{arg\,min}

\DeclareMathOperator{\Tr}{Tr}

\usepackage{authblk}

\title{Ego-based Entropy Measures for Structural Representations}

\author[1,2]{George Dasoulas}
\author[1]{Giannis Nikolentzos} 
\author[2]{Kevin Scaman}
\author[2]{Aladin Virmaux}
\author[1]{Michalis Vazirgiannis}
\affil[1]{\'Ecole Polytechnique}
\affil[2]{Noah’s Ark Lab, Huawei}
\begin{document}

\maketitle
\begin{abstract}
In complex networks, nodes that share similar structural characteristics often exhibit similar roles (e.g type of users in a social network or the hierarchical position of employees in a company). In order to leverage this relationship, a growing literature proposed latent representations that identify structurally equivalent nodes. However, most of the existing methods require high time and space complexity. In this paper, we propose \MN, a simple approach for generating low-dimensional structural node embeddings, that is both time efficient and robust to perturbations of the graph structure. The proposed approach focuses on the local neighborhood of each node and employs the Von Neumann entropy, an information-theoretic tool, to extract features that capture the neighborhood's topology.  Moreover, on graph classification tasks, we suggest the utilization of the generated structural embeddings for the transformation of an attributed graph structure into a set of augmented node attributes. Empirically, we observe that the proposed approach exhibits robustness on structural role identification tasks and state-of-the-art performance on graph classification tasks, while maintaining very high computational speed.
\end{abstract}

\section{Introduction}

The amount of data that can be represented as graphs has increased significantly in recent years. Graph representations are ubiquitous in several fields such as in biology, chemistry and social networks~\citep{hamilton}.
Many applications require performing machine learning tasks on such type of data.
For instance, in chemistry, graph regression can successfully replace expensive quantum mechanical simulation approaches in predicting the quantum properties of organic molecules~\citep{pmlr-v70-gilmer17a}. 

The past few years have witnessed great activity in the field of learning on graphs. This activity has led to the development of several sophisticated approaches. In the supervised setting, graph neural networks have achieved great success in tackling both node and graph-related problems~\citep{Scarselli:2009:GNN:1657477.1657482,Kipf:2016tc}. In the unsupervised setting, most of the activity has focused on node embedding algorithms~\citep{hamilton, DBLP:journals/corr/abs-1806-08804}. So far, most of these algorithms are designed so that they preserve the proximity between nodes, i.e., nodes that are close to each other in the graph (or belong to the same community) obtain similar representations, while distant nodes are assigned completely different representations~\citep{perozzi2014deepwalk,Grover:2016:NSF:2939672.2939754}. However, some tasks require assigning similar embeddings to nodes that perform similar functions in the network, regardless of their distance.
These tasks require \emph{structural embeddings}, i.e. embeddings that can identify structural properties of a node's neighborhood.
For instance, predicting job positions based on the communication network of a company may be achieved by observing the type of interactions that user have with their colleagues (e.g. broadcasting to a large audience for secretaries or frequent communication to a small group for teammates).

\section{Related Work}
The major part of the research interest in the field of node embeddings focuses on encouraging nodes close to each other to have similar representations, based on the homophily conditions that satisfy many social and bioinformatics networks. In this work, we focus on the structural equivalence between nodes rather than the homophily, in order to provide structural representations. RolX algorithm~\citep{rolx} is one of the first and still successful approaches to model structural node roles. Specifically, it extracts features for each node and applies non-negative matrix factorization to the emerging matrix in order to automatically discover node roles in the graph. A more recent approach, struc2vec~\citep{DBLP:journals/corr/FigueiredoRS17}, constructs a multi-layer graph which encapsulates structural characteristics of the original graph. It then performs random walks to learn structural representations.
GraphWave~\citep{donat} is another approach which capitalizes on the eigenspectrum of the graph to compute diffusion wavelets in the complex space, and uses the real and imaginary parts of these wavelets to generate node embeddings. One main drawback of this method is that it suffers from high space complexity when applied to large graphs. DRNE~\citep{drne} aggregates neighborhood information using an LSTM operator upon the sequence of a node's neighbors. This method implies an ordering of each node's neighborhood, hence not satisfying the permutation invariance criteria and, thus, could affect its performance.  Regarding the utilization of structural characteristics in order to decompose the graph structure and the attribute space of the graph nodes, a recent work~\citep{DBLP:journals/corr/abs-1905-04579} proposes the augmentation of the node attribute vectors with features that encode the graph structure. 

\paragraph{Contribution.}
In this paper, we provide a novel and simple structural node embedding algorithm which capitalizes on information-theoretic tools. The algorithm employs the Von Neumann entropy to construct node representations related to the structural identity of the neighborhood of each node. These representations capture the structural symmetries of the neighborhoods of increasing radius of each node. The algorithm is evaluated in node classification and node clustering tasks where it achieves performance comparable to state-of-the-art methods.
Moreover, the algorithm is evaluated on standard graph classification datasets where it outperforms recently-proposed graph neural network models. Code will be available at \href{https://github.com/}{https://github.com/} after the review process.

\begin{figure*}[htp]
    \centering
    \includegraphics[width=\textwidth,height=3.8cm]{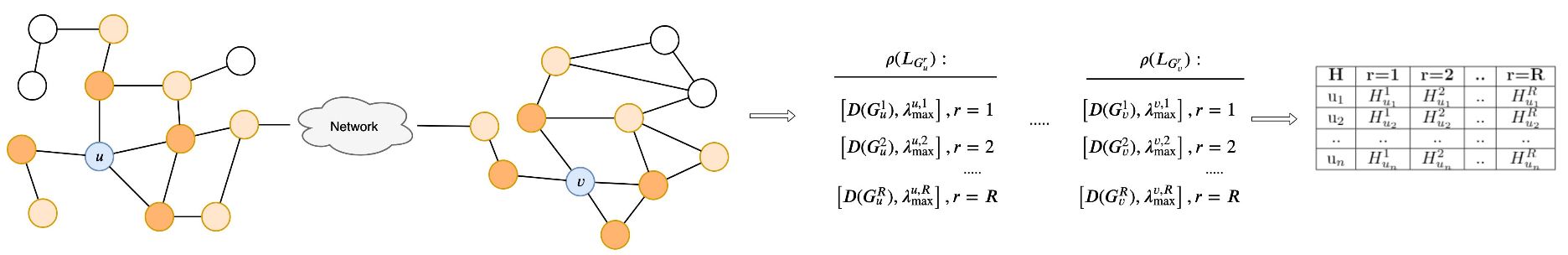}
    \caption{\MN extracts ego-networks for a series of defined radii and computes the Von Neumann entropy for every radius. On the left, the two parts of the network may have a large distance through the network. The 1-hop ego-networks are highlighted with the dark yellow, while the remaining nodes of the 2-hop ego-networks are highlighted with the light yellow. The nodes $u,v$ have structurally equivalent 1-hop neighborhoods $G_u^1$ and $G_v^1$, though their 2-hop neighborhoods $G_u^2$, $G_v^2$ hold different structural characteristics.}
    \label{fig:method}
\end{figure*}

\section{Structural Embeddings based on Von Neumann Entropy}

\vspace{1mm}
We next present the proposed approach for generating structural node embeddings, employing the Von Neumann entropy, a model-agnostic measure, that quantifies the structural complexity of a graph. Graph entropy methods have been used in recent works~\citep{lipan, shetty} for graph similarity in network analysis problems. The Von Neumann graph entropy (VNGE) has been proven to have a linear correlation with other graph entropy measures~\citep{anand}. Based on its applications, our method exploits VNGE as a similarity measure between neighborhoods of nodes, in order to extract structural representations. 

\subsection{Von Neumann Entropy on Graphs}
\vspace{1mm}
In the field of quantum mechanics, the state of a quantum mechanical system is described by a \emph{density} matrix $\rho$, i.e a positive semidefinite, hermitian matrix with unit trace~\citep{neumann}.
Given the above, the Von Neumann entropy of the quantum system is defined as: \begin{equation}
    \label{eq::1}
    H(\rho) = -\Tr(\rho\log\rho) = -\sum_{i=1}^n \lambda_i\log\lambda_i,
\end{equation}
where $\Tr(\cdot)$ is the trace of a matrix, and $\lambda_i$'s are the eigenvalues of $\rho$. Correspondigly, given a graph $G=(V,E)$, where $|V|=n$ and its laplacian $L_G= D-A$, where $D$ is the degree matrix and $A$ the adjacency matrix, the scaled matrix $\rho(L_G) = \frac{L_G}{\text{Tr}(L_G)} = \frac{L_G}{2|E|}$ is symmetric, positive semidefinite and with unit trace, suggesting an analogy with the density matrix~\citep{Braunstein2006}. Thus, the Von Neumann graph entropy (VNGE) is defined as:
$H \big(\rho(L_G) \big) = -\Tr \big(\rho(L_G)*\log\rho(L_G) \big)$.
Note that $\lambda_i = \frac{1}{\text{Tr}(L_G)}v_i$ where $v_i$ is the $i$-th eigenvalue of $L_G$. Therefore, $0 \leq \lambda_i \leq 1$ holds for all $i \in \{ 0,1,\ldots,n \}$~\citep{Passerini2009QuantifyingCI}.
This indicates that \Eqref{eq::1} is equivalent to the Shannon entropy of the probability distribution $\{\lambda_i\}_{i=1}^n$.
Hence, $H \big( \rho(L_G) \big)$ serves as a measure of skewness of the eigenvalue distribution and it has been shown that it provides information about the spectral complexity of a graph and that is related to different structural characteristics of the graph~\citep{Passerini2009QuantifyingCI}.

\paragraph{Efficient approximation scheme.}
The computation of the VNGE requires performing the eigenvalue decomposition of the density matrix which can be done in $\mathcal{O}(n^3)$ time. That means that, in cases of large graphs, the complexity of this computation is very high.  
Recent works~\citep{fast_incr, choi2018fast} have proposed an efficient approximation of $H(\rho)$.
They first compute a quadratic approximation of its value, and then combine it with the largest eigenvalue. In particular, starting from \Eqref{eq::1} and following~\citep{minello}, we obtain:
\begin{equation}\label{eq::3}
    H(\rho) = -\text{Tr}(\rho\log\rho) \approx \text{Tr}\big(\rho(I_n - \rho)\big) = Q\,,
\end{equation}
where $I_n$ is the $n \times n$ identity matrix, and
\begin{equation}\label{eq::4}
    \begin{split}
        Q &= \frac{1}{2m}\times\text{Tr}(L_G) - \frac{1}{4m^2}\times\text{Tr}(L_G^2)\\
        &= 1 - \frac{1}{2m} - \frac{1}{4m^2}\sum_{i=1}^n d_i^2\,,
    \end{split}
\end{equation}
where $m=|E|$ is the number of edges of the graph, and $d_i$ is the degree of the $i$-th node.
Note that $Q$ corresponds to the quadratic approximation of $H(\rho)$.
Finally, as~\citep{fast_incr} suggest, we can obtain a tighter approximation of $H(\rho)$ as follows:
\begin{equation}\label{eq::5}
 \hat{H} = -Q\ln\lambda_{\max}\,,
\end{equation}
where $\lambda_{\max}$ is the largest eigenvalue of the Laplacian $L(G)$.
It can be shown that for any density matrix $\rho$, we have $H(\rho) \geq \hat{H}(\rho)$ where the equality holds if and only if $\lambda_{\max} = 1$~\citep{choi2018fast}.

\subsection{The \MN Algorithm}
\vspace{1mm}
Next, based on the VNGE and its approximation, we introduce our proposed approach, in order to construct structural representations. The \MN algorithm extracts ego-networks of increasing radius and computes their VNGE.
Then, the representation of a node comprises of the Von Neumann entropies that emerged from the node's ego-networks.
Therefore, the set of entropies of the ego-networks of a node serve as a ``signature'' of the structural identity of its neighborhood.

Let $R$ be the maximum considered radius. For each $r \in \{1,..,R\}$ and each node $v \in V$, the algorithm extracts the $r$-hop neighborhood  $G_v^r = (V',E')$, and the Laplacian $L_{G_v^r}$, where $V' = \{u \in V | d(u,v) \leq r\}$ and $E' = \{(u,v) | u, v \in V', (u,v) \in E \}$.
Next, the algorithm computes the density matrix $\rho(L_{G_v^r})$ and its eigenvalues. $H \big(\rho(L_{G_v^r})\big)$ of the $r$-hop neighborhood of $v$ is computed using \Eqref{eq::5}.
Finally, the $R$ entropies are arranged into a single vector (i.e., node embedding) $h_v \in \mathbb{R}^R$.
The method is illustrated in Algorithm~\ref{algo:vne} below. 

As illustrated in Figure~\ref{fig:method}, \MN is able to identify structural equivalences between nodes, which are distant to each other. Specifically, nodes $u$ and $v$ share structurally identical $1$-hop neighborhoods.
Therefore, the entropies of their $1$-hop neighborhoods are equal to each other.
However, this is not the case for the entropies of their $2$-hop neighborhoods since the two subgraphs are very dissimilar from each other. Note that in this work, we focus on undirected graphs without edge weights.  However, our approach easily extends in the case of weighted and directed graphs.

\makeatletter
\newcommand{\removelatexerror}{\let\@latex@error\@gobble}
\makeatother

\begingroup
\removelatexerror
\begin{algorithm}[H]
\SetAlgoLined
\SetKwInOut{Input}{input}\SetKwInOut{Output}{output}
\SetKwFor{ForEach}{foreach}{do}{endfor}
 
\Input{graph $G=(V,E)$, radius $R$}
\Output{An embedding matrix $H \in \mathbb{R}^{n\times R}$ }
 \ForEach{$v \in V$}{
    \ForEach{$r \in \{1,\ldots,R\}$}{
    Extract ego-graph Laplacian $L_{G_v^r}$\\
    Derive density matrix $\rho(L_{G_v^r}) = \frac{L_{G_v^r}}{\text{Tr}(L_{G_v^r})}$\\
    $ \lambda_{\max} \leftarrow \text{Power-Iteration}(\rho(L_{G_v^r}))$\\
    $Q \leftarrow 1 - \frac{1}{2m} - \frac{1}{4m^2}\sum_{i=1}^{|V'|} d_i^2$
    
    $h(v,r) \leftarrow -Q \ln\lambda_{\max}$ 
    }
    $H_v \leftarrow$ CONCAT$(h(v,1),...,h(v,R))$
 }
 \caption{\MN algorithm}
 \label{algo:vne}
\end{algorithm}
\endgroup

\paragraph{Computational Complexity.}
Algorithm~\ref{algo:vne} consists of two computational steps: ($1$) the extraction of the ego-networks and ($2$) the computation of VNGEs for all subgraphs.
The first step is linear in the number of edges of the node's neighborhood.
In the worst case, the complexity is $\mathcal{O}(nm)$, but for sparse graphs and for small values of $R$, the complexity is significantly lower (constant in practice).
With regards to the second step, as we mentioned above, we do not perform the eigenvalue decomposition of the \emph{density} matrix (requires $\mathcal{O}(n^3)$ time), but we make use of \Eqref{eq::5}.
This requires finding the largest eigenvalue $\lambda_{max}$ and computing simple degree statistics for each $r$-hop neighborhood.  
We use the power iteration method~\citep{power_iteration} to compute $\lambda_{max}$, which requires $\mathcal{O}(n+m)$ operations, as the Laplacian matrix has $n+m$ nonzero entries.
Hence, the whole approximation exhibits linear complexity $\mathcal{O}(n+m)$, while for very sparse graph, it becomes $\mathcal{O}(n)$.

\paragraph{Robustness over "small" perturbations.}
We will next show that utilizing the VNGE, we can acquire robust structural representations over possible perturbations on the graph structure. Clearly, if two graphs are isomorphic to each other, then their entropies will be equal to each other. It is important, though, for structurally similar graphs to have similar entropies, too. So, let $\rho, \rho' \in \mathbb{R}^{n \times n}$ be the density matrices of two graph laplacians $L_G,L_{G'}$, as described above.
Let also $\rho = P \rho' P^\top + \epsilon$ where $P$ is an $n \times n$ permutation matrix equal to $\argmin_{P} || \rho - P \rho' P^\top ||_F$ and $\epsilon$ is an $n \times n$ symmetric matrix.
If $G,G'$ are nearly-isomorphic, then the Frobenius norm of $\epsilon$ is small. By applying the Fannes-Audenaert inequality~\citep{Audenaert_2007}, we have that:
\begin{equation*}
 |H(\tilde{\rho}) - H(\rho)| \leq \frac{1}{2}T \ln(n-1) + S(T),   
\end{equation*} where $T = ||\tilde{\rho}-\rho||_1$ is the trace distance between $\rho,\tilde{\rho}$ and $S(T) = -T\log T- (1-T)\log(1-T)$.
However, $||\tilde{\rho}-\rho||_1 = \sum_i|\lambda_i^{\tilde{\rho}-\rho}| \leq n||\tilde{\rho} - \rho ||_{op}$, where $|| \cdot ||_{op}$ is the operator norm.
Therefore, $|H(\tilde{\rho})- H(\rho)| \leq \frac{N}{2}ln(N-1)||\epsilon||_{op} + S[\{T,1-T\}]$, leading thus to an upper bound of the difference between the entropies of structurally similar graphs.

\vspace{3mm}
\subsection{Graph-level Representations}
Next, we propose incorporating the structural embeddings generated by \MN into graph classification algorithms.
The majority of the state-of-the-art methods learn node representations using message-passing schemes~\citep{hamilton, xu2018powerful}, where each node updates its representation $T$ times by aggregating the representations of its neighbors and combining them with its own representation.
Clearly, each time the nodes update their representations, the structure of the graph is taken into account.
In this work, we do not use any message-passing scheme and we ignore the graph structure.
Instead, we employ the \MN and we embed the nodes into a low-dimensional space.
These embeddings are then combined with the node attributes (if any).
In fact, information about the graph structure is incorporated into the embeddings generated by the proposed algorithm.
This approach follows recent studies that propose to augment the node attribute vectors with structural characteristics, in order to avoid performing some message-passing procedure~\citep{DBLP:journals/corr/abs-1905-04579, errica2019fair}.
The above pre-processing step transforms the graph into a set of vectors (i.e., one vector for each node).
Then, these vectors are passed on to a neural network model which transforms them and then aggregates them using some permutation invariant function~\citep{NIPS2017_6931}.
Specifically, given a matrix of node attributes $X \in \mathbb{R}^{n\times d}$, our approach performs the following steps:
\begin{itemize}
    \itemsep0em 
    \item Computation of $H_v \in \mathbb{R}^{n\times R}$
    \item Concatenation of node attribute vectors with structural node embeddings: $X' = [X || H] \in \mathbb{R}^{n\times (d+R)}$
    \item Aggregation of node vectors $X'$ into graph embedding $H_G = \psi( \sum_{v \in V_G} \phi(X'_v)) $, where $\phi$ and $\psi$ are neural networks.
\end{itemize}
The above procedure does not apply any message-passing scheme.
This reduces the computational complexity of the training procedure since each graph is represented as a sets of node representations.

\section{Experiments}
Next, we evaluate the performance of the proposed approach in two scenarios: (1) the structural role identification task, where we extract the role of a node in the graph and (2) the graph classification task, where given a graph (attributed or not), we predict the class that it belongs, based on the graph-level representation. For the structural role identification task we use both synthetic and real-world graphs, while for the graph classification task we use 5 well-studied real-world datasets.

\subsection{Structural Role Identification}
We first experiment with some synthetic datasets and then we compare the performance of \MN and baselines on a real-world dataset.
\subsubsection{Toy network: Barbell Graph}

\begin{figure}[h!]
    \centering
        \includegraphics[width=\columnwidth, height =2.6cm]{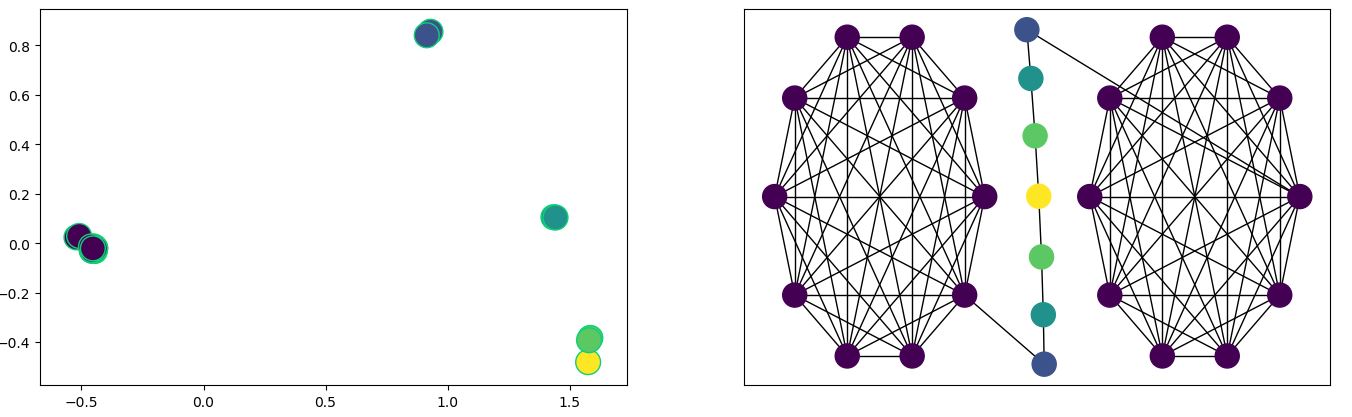}    
    \caption{\small{The barbell graph (right) and the $2$-dimensional representations of its nodes produced by applying PCA to the \MN embeddings (left).}}
    \label{fig:barbell}
\end{figure}

This toy graph consists of two cliques of size $10$ that are connected through a path of length $7$.
The graph is shown in \Figref{fig:barbell} (right). The different colors indicate the roles of the nodes in the graph.
\Figref{fig:barbell} (left) illustrates the $2$-dimensional representations of the $27$ nodes of the graph.
These representations were generated by the \MN algorithm (we set $R=3$ and then applied PCA to project them to the $2$-dimensional space).
We should mention that the proposed algorithm can identify the structural role of the nodes in the barbell graph and produce similar/identical embeddings for structurally similar/identical nodes.

\subsubsection{Highly-symmetrical synthetic networks}
In order to evaluate the expressiveness of the structural embeddings generated by our method, we measure its performance on synthetic datasets, which were introduced in~\citep{donat, DBLP:journals/corr/FigueiredoRS17}.
We perform both classification and clustering with the same experimentation setup as~\citep{donat}.

\paragraph{Evaluation.}
For the classification task, we measure the \textit{accuracy} and the \textit{F1-score}. 
For the clustering task, we report the $3$ evaluation metrics, that were also calculated in~\citep{donat}: \textit{Homogeneity}, \textit{Completeness} and \textit{Silhouette}. Specifically, the \textit{homogeneity} evaluates the conditional entropy of the structural roles in the generated clustering, based on each method: $H(C) = - \sum_{c=1}^{|C|} \frac{\sum_{k=1}^{|K|} a_{ck}}{n} \log \frac{\sum_{k=1}^{|K|} a_{ck}}{n} $ where $C$ is the set of the different classes-structural roles $C=  \{c_1,c_2,..,c_{|C|}\}$, $K$ is the set of the assigned clusters $K=  \{k_1,k_2,..,k_{|K|}\}$ and $a_{ij}$ is the number of nodes with structural role $c$ and assigned in the cluster $j$. The \textit{completeness} evaluates how many nodes with equivalent structural roles are assigned to the same cluster: $H(C) = - \sum_{k=1}^{|K|} \frac{\sum_{c=1}^{|C|} a_{ck}}{n} \log \frac{\sum_{c=1}^{|C|} a_{ck}}{n}$. The \textit{silhouette} measures the mean intra-cluster distance vs the mean inter-cluster distance. Regarding our method, the only hyperparameter that we optimized was the radius of the considered ego-networks. We chose $R$ from $\{1,..,3\}$.
\paragraph{Dataset setup.}
The generated synthetic datasets are identical to those used in~\citep{donat}. They consist of basic symmetrical shapes, as shown in Table~\ref{tab:shapes}, that are regularly placed along a cycle of length $30$. The \textit{basic} setups use 10 instances of only one of the shapes of Table~\ref{tab:shapes}, while the \textit{varied} setups use 10 instances of every shape, randomly placed along the cycle. 
The perturbed instances are formed by randomly rewiring edges.
The colors in the shapes indicate the different classes.

\begin{table*}[h!]
\centering
\def\arraystretch{1.1}
\resizebox{\textwidth}{!}{
\begin{tabular}{|lc|l||ccc|cc|} \hline
Configuration & Shapes & Algorithm & Homogeneity & Completeness & Silhouette & Accuracy & F$1$-score \\ \hline
\multirow{7}{*}{Basic} & \multirow{7}{*}{\includegraphics[width=.2\textwidth]{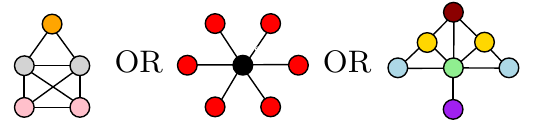}} & DeepWalk & 0.178 & 0.115 & 0.163 & 0.442 & 0.295 \\
& & RolX& \textbf{0.983} & \textbf{0.976} & 0.846 & \textbf{1.000} & \textbf{1.000} \\
& & struc2vec& 0.803 & 0.595 & 0.402 & 0.784 & 0.708 \\
& & GraphWave& 0.868 & 0.797 & 0.730 & 0.995 & 0.993 \\
& & VNEstruct& 0.966 & 0.963 & \textbf{0.891} & 0.920 & 0.901 \\ \hline
\multirow{7}{*}{\parbox{1cm}{Basic\\ Perturbed}} & \multirow{7}{*}{\includegraphics[width=.2\textwidth]{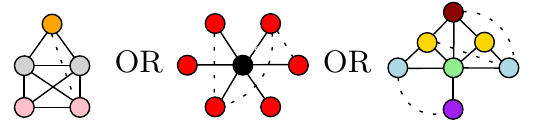}} & DeepWalk& 0.172 & 0.124 & 0.171 & 0.488 & 0.327 \\
& & RolX& 0.764 & 0.458 & 0.429 & 0.928 & \textbf{0.886} \\
& & struc2vec& 0.625 & 0.543 & 0.424 & 0.703 & 0.632 \\
& & GraphWave& 0.714 & 0.326 & 0.287 & 0.906 & 0.861 \\ 
& & VNEstruct& \textbf{0.882} & \textbf{0.701} & \textbf{0.478} & \textbf{0.940} & 0.881 \\ \hline 
\multirow{7}{*}{Varied} &
\multirow{7}{*}{\includegraphics[width=.2\textwidth]{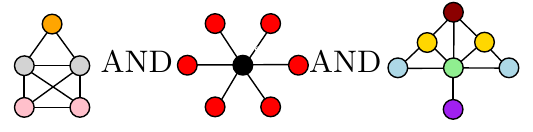}} & DeepWalk& 0.327 & 0.220 & 0.216 & 0.329 & 0.139 \\
& & RolX& \textbf{0.984}  & \textbf{0.939} & 0.748 & \textbf{0.998} & 0.996 \\
& & struc2vec& 0.805 & 0.626 & 0.422 & 0.738 & 0.592 \\
& & GraphWave& 0.941 & 0.843 & \textbf{0.756} & 0.982 & \textbf{0.965} \\ 
& & VNEstruct& 0.950  & 0.892 & 0.730 & 0.988 & 0.95 \\ \hline 
\multirow{7}{*}{\parbox{1cm}{Varied\\ Perturbed}} & \multirow{7}{*}{\includegraphics[width=.2\textwidth]{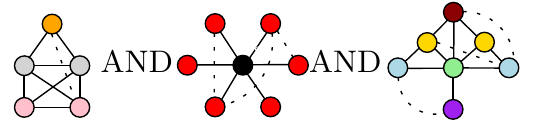}} & DeepWalk& 0.300 & 0.231 & 0.221 & 0.313 & 0.128 \\
& & RolX& 0.682 & 0.239 & 0.062 & 0.856 & 0.768 \\
& & struc2vec& 0.643 & 0.524 & \textbf{0.433} & 0.573 & 0.412 \\
& & GraphWave& 0.670 & 0.198 & 0.005 & 0.793 & 0.682 \\ 
& & VNEstruct& \textbf{0.722} & \textbf{0.678} & 0.399 & \textbf{0.899}  & \textbf{0.878}  \\ \hline 
\end{tabular}
}
\caption{Performance of the baselines and the \textit{VNEstruct} method for learning structural embeddings averaged over $20$ synthetically generated graphs. Dashed lines denote perturbed graphs.}
\label{tab:shapes}
\end{table*}

\begin{figure}[h!]
    \centering
    \includegraphics[scale=0.72]{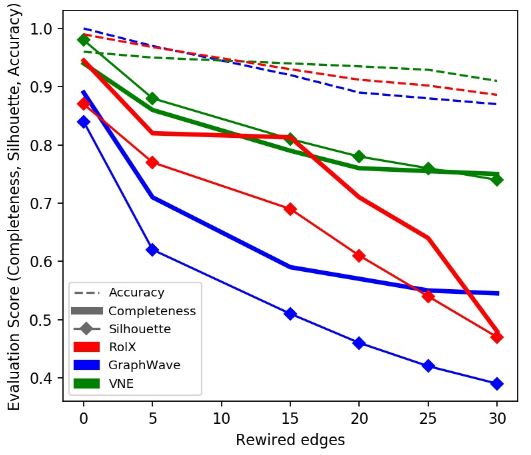}
    \caption{Classification and clustering performance of \MN and the baselines with respect to noise.}
    \label{fig:performances}
\end{figure}

As Table~\ref{tab:shapes} shows, VNEstruct outperforms the competitors on the perturbed instances of the synthetic graphs.
Specifically, while on the basic and the varied configurations RolX and GraphWave achieve higher F1-scores, on the perturbed configurations GraphWave and VNEstruct show better performance, with our approach outperforming all the others in the varied perturbed configuration.
The results in Table~\ref{tab:shapes} suggest a comparison of VNEstruct, RolX and GraphWave on noisy setting. This comparison is provided in \Figref{fig:performances}.
Here, we use the same dataset, but we report the classification and clustering performance with respect to the number of rewired edges (from $0$ to $20$).
As we can see, the presence of noise has a less impact on VNEstruct than on GraphWave and RolX, especially in the clustering task.

\subsubsection{Role identification on Email-dataset}
We next evaluate the performance of VNEstruct algorithm and of its competitors on a real-world dataset, the Enron Email dataset~\citep{enron}.
This is an email network, where nodes are company employees with one of seven different positions (e.g. CEO, Manager, Employee etc.) and the edges indicate the email communication. It contains 143 nodes and 2,583 edges. We consider that nodes with structurally similar positions in the graph share similar roles. Following the same experimental protocol as in~\citep{donat}, we generate  structural embeddings.
Regarding the \MN, we hyperparameterized over the radius $r \in \{1,2,3\}$, since the graph diameter is 4.
Table~\ref{tab::enron} shows that \MN  with  radius $R=3$ and a low embedding dimensionality ($d\leq 3$) achieves competitive results with strong baselines, which produce embeddings with higher dimensionality ($d>10$).

\begin{table}[htp]
\centering
\begin{tabular}{|l|cccc|}
\hline
 Method    & C  & H    & S & Dimension    \\ \hline
 RolX      & 0.028          & 0.090          & 0.425 & 16 - 32          \\ 
 struc2vec & 0.018          & 0.003          & 0.435 & 64 - 128         \\ 
 GraphWave & \textbf{0.067} & \textbf{0.115} & 0.577 & 50 - 100         \\ 
 \MN & 0.049          & 0.107          & \textbf{0.591} & 1 - 3 \\ \hline
\end{tabular}

\caption{Clustering evaluation for Enron Email. C, H, S stand for Completeness, Homogeneity and Silhouette.}
\label{tab::enron}
\end{table}

\begin{table*}[htp]
\centering
\begin{tabular}{|ccccc|}
\hline
  Method    & MUTAG   & IMDB-BINARY    & PTC-MR & PROTEINS     \\ \hline
  DGCNN      &  85.83 $\pm$ 1.66 & 70.03 $\pm$ 0.86 & 58.62$\pm$2.34 & 75.54 $\pm$ 0.94 \\ 
 
  CapsGNN      & 86.67 $\pm$ 6.88          & 73.10 $\pm$ 4.83         & - & 76.28 $\pm$ 3.63 \\ 
  GIN & 89.40 $\pm$ 5.60 & 75.10 $\pm $ 5.10          & 64.6 $\pm$ 7.03 & 76.20 $\pm$ 2.60  \\ 
   GCN & 87.20 $\pm$ 5.11  & 73.30 $\pm$ 5.29 & 64.20 $\pm$ 4.30  & 75.65 $\pm$ 3.24 \\ 
\hline
  GFN & 90.84 $\pm$ 7.22 & 73.00 $\pm$ 4.29 & - &  \textbf{77.44 $\pm$ 3.77}\\ 
  \MN & \textbf{91.08 $\pm$ 5.65 }         & \textbf{75.40 $\pm$ 3.33}          & \textbf{65.39 $\pm$ 8.57}  & 77.41 $\pm$ 3.47  \\ 
  \hline
\end{tabular}
\caption{Average classification accuracy ($\pm$ standard deviation) of the baselines and the proposed \MN algorithm on the $5$ graph classification datasets.}

\label{tab::graphclass}
\end{table*}

\subsection{Graph Classification}
\vspace{1mm}
\subsubsection{Graph classification on molecular and social networks}

\vspace{1mm}
We next evaluate \MN algorithm and the baselines in the task of graph classification.
We compare our proposed algorithm against well-established message-passing algorithms for learning graph representations.
Note that in contrast to the majority of the baselines, we pre-compute the entropy-based structural representations, and then we represent each graph as a set of vectors (i.e., its node repesentations) which encode structural charasteristics of the neighborhood of each node.

\paragraph{Datasets.}
We use 4 graph classification datasets (3 are from bionformatics: MUTAG, PROTEINS, PTC-MR and 1 dataset comes from social-networks: IMDB-BINARY). The datasets have been examined in a variety of graph kernels and graph neural networks methods~\citep{xu2018powerful, Shervashidze:2011:WGK:1953048.2078187, Kipf:2016tc}. As they have been previously described, the bioinformatics datasets contain node attributes, while the social network does not and following previous works, we create the attributes by employing one-hot encodings of the node degrees. In the case of \MN, we append to the attributes the generated structural embeddings.
\paragraph{Baselines.}
The goal of the comparison is to show that decomposing the graph structure and the attribute space, we can achieve comparable results to the state-of-the-art algorithms. Thus, we use as baselines graph neural network variants and specifically: DGCNN~\citep{Zhang2018AnED}, Capsule GNN~\citep{Xinyi2019CapsuleGN}, Graph Isomorphism Network~\citep{xu2018powerful}, Graph Convolutional Network~\citep{Kipf:2016tc}. Moreover, in a more recent work~\citep{DBLP:journals/corr/abs-1905-04579}, the authors propose Graph Feature Network, which, also, augments the attributes with structural features and then ignores the graph structure during the learning procedure. 
\paragraph{Model setup.}
For the baselines, we followed the same experimentation setup, as described on~\citep{DBLP:journals/corr/abs-1905-04579} and, thus, we report the achieved accuracies. Regarding the \MN, we performed 10-fold cross-validation with Adam optimizer and learning rate decay every 50 epochs by a factor of 0.3. In all experiments, we set the number of epochs to 300. As hyper-parameters , we set the radius of the ego-networks $r \in \{1,2,3,4\}$ and the number of hidden layers $d \in \{8,16,32\}$ on the MLPs of the node representation aggregator.

\begin{figure}[htp]
    \centering
    \includegraphics[width=\columnwidth, height=4.5cm]{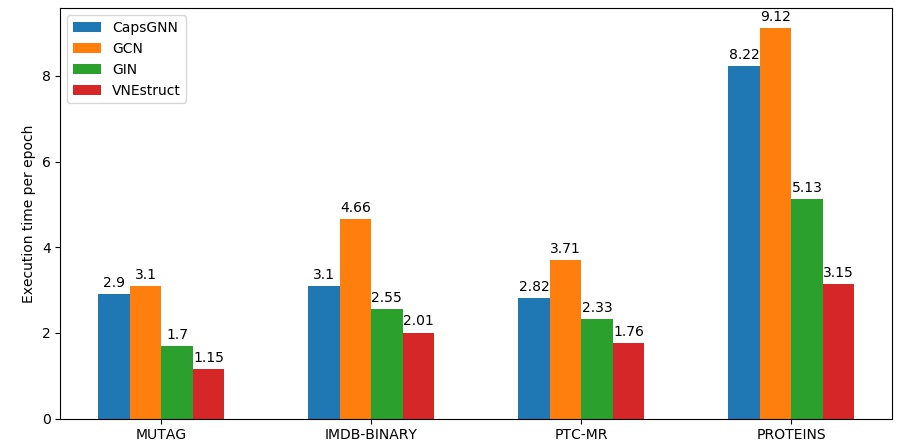}
    \caption{Training time per epoch (in sec) of VNEstruct and competitors for the graph classification tasks. }
    \label{fig:speed}
\end{figure}

\paragraph{Performance and efficiency results.}

Table~\ref{tab::graphclass} illustrates the average classification accuracies of the proposed approach and the baselines on the $5$ graph classification datasets.
Interestingly, the proposed approach achieves accuracies comparable to some of the state-of-the-art message-passing models. \MN outperformed all the baselines on $3$ out of $4$ datasets, while achieved the second best accuracy on the remaining dataset PROTEINS.

With regards to the running time of the different methods, Figure~\ref{fig:speed} illustrates the average training time per epoch of
VNEstruct and some baselines that apply message-passing schemes.
The proposed approach is generally more efficient than the baselines.
Specifically, it is $0.31$ times faster than GIN and $0.60$ times faster than GCN on average.
This improvement in efficiency is mainly due to the fact that the graph structural features are computed in a preprocessing step, are then concatenated with the node attributes, and are passed on the neural network model.
Furthermore, we should mention that due to the low dimensionality of the generated embeddings ($d \leq 4$), our method does not have any significant requirements in terms of memory.

\section{Conclusion}

In this paper, we proposed an algorithm for generating structural node representations, based on the entropies of ego-networks. We evaluated the proposed algorithm in node classification and clustering tasks where it either outperformed or performed comparably to strong baselines. 
We also proposed an approach for performing graph-related tasks which combines these representations with the nodes' attributes, and then passes the new representations into a neural network model, avoiding the computational cost of message passing schemes. 
The proposed approach yielded high classification accuracies on standard datasets.


\bibliography{biblio}

\begin{thebibliography}{}

\bibitem[\protect\citeauthoryear{Anand \bgroup \em et al.\egroup
  }{2011}]{anand}
Kartik Anand, Ginestra Bianconi, and Simone Severini.
\newblock Shannon and von neumann entropy of random networks with heterogeneous
  expected degree.
\newblock {\em Phys. Rev. E}, 83:036109, Mar 2011.

\bibitem[\protect\citeauthoryear{Audenaert}{2007}]{Audenaert_2007}
Koenraad M~R Audenaert.
\newblock A sharp continuity estimate for the von neumann entropy.
\newblock {\em Journal of Physics A: Mathematical and Theoretical},
  40(28):8127–8136, Jun 2007.

\bibitem[\protect\citeauthoryear{Braunstein \bgroup \em et al.\egroup
  }{2006}]{Braunstein2006}
Samuel~L. Braunstein, Sibasish Ghosh, and Simone Severini.
\newblock The laplacian of a graph as a density matrix: A basic combinatorial
  approach to separability of mixed states.
\newblock {\em Annals of Combinatorics}, 10(3):291--317, Dec 2006.

\bibitem[\protect\citeauthoryear{Chen \bgroup \em et al.\egroup
  }{2019a}]{fast_incr}
Pin-Yu Chen, Lingfei Wu, Sijia Liu, and Indika Rajapakse.
\newblock Fast incremental von neumann graph entropy computation: Theory,
  algorithm, and applications.
\newblock In {\em International Conference on Machine Learning (ICML2019), June
  2019}, ICML '19, 2019.

\bibitem[\protect\citeauthoryear{Chen \bgroup \em et al.\egroup
  }{2019b}]{DBLP:journals/corr/abs-1905-04579}
Ting Chen, Song Bian, and Yizhou Sun.
\newblock Are powerful graph neural nets necessary? {A} dissection on graph
  classification.
\newblock {\em CoRR}, abs/1905.04579, 2019.

\bibitem[\protect\citeauthoryear{Choi \bgroup \em et al.\egroup
  }{2018}]{choi2018fast}
Hayoung Choi, Jinglian He, Hang Hu, and Yuanming Shi.
\newblock Fast computation of von neumann entropy for large-scale graphs via
  quadratic approximations, 2018.

\bibitem[\protect\citeauthoryear{Donnat \bgroup \em et al.\egroup
  }{2018}]{donat}
Claire Donnat, Marinka Zitnik, David Hallac, and Jure Leskovec.
\newblock Learning structural node embeddings via diffusion wavelets.
\newblock In {\em Proceedings of the 24th ACM SIGKDD International Conference
  on Knowledge Discovery $\&$ Data Mining}, KDD '18, pages 1320--1329, New
  York, NY, USA, 2018. ACM.

\bibitem[\protect\citeauthoryear{Errica \bgroup \em et al.\egroup
  }{2019}]{errica2019fair}
Federico Errica, Marco Podda, Davide Bacciu, and Alessio Micheli.
\newblock A fair comparison of graph neural networks for graph classification,
  2019.

\bibitem[\protect\citeauthoryear{Gasser}{1970}]{neumann}
W.~Gasser.
\newblock J. v. neumann, mathematische grundlagen der quantenmechanik.
\newblock {\em ZAMM - Journal of Applied Mathematics and Mechanics /
  Zeitschrift für Angewandte Mathematik und Mechanik}, 50(6):437--438, 1970.

\bibitem[\protect\citeauthoryear{Gilmer \bgroup \em et al.\egroup
  }{2017}]{pmlr-v70-gilmer17a}
Justin Gilmer, Samuel~S. Schoenholz, Patrick~F. Riley, Oriol Vinyals, and
  George~E. Dahl.
\newblock Neural message passing for quantum chemistry.
\newblock In {\em Proceedings of the 34th International Conference on Machine
  Learning}, pages 1263--1272, 2017.

\bibitem[\protect\citeauthoryear{Grover and
  Leskovec}{2016}]{Grover:2016:NSF:2939672.2939754}
Aditya Grover and Jure Leskovec.
\newblock Node2vec: Scalable feature learning for networks.
\newblock In {\em Proceedings of the 22Nd ACM SIGKDD International Conference
  on Knowledge Discovery and Data Mining}, KDD '16, pages 855--864, New York,
  NY, USA, 2016. ACM.

\bibitem[\protect\citeauthoryear{Hamilton \bgroup \em et al.\egroup
  }{2017}]{hamilton}
William~L. Hamilton, Rex Ying, and Jure Leskovec.
\newblock Representation learning on graphs: Methods and applications.
\newblock {\em CoRR}, abs/1709.05584, 2017.

\bibitem[\protect\citeauthoryear{Henderson \bgroup \em et al.\egroup
  }{2012}]{rolx}
Keith Henderson, Brian Gallagher, Tina Eliassi-Rad, Hanghang Tong, Sugato Basu,
  Leman Akoglu, Danai Koutra, Christos Faloutsos, and Lei Li.
\newblock Rolx: Structural role extraction $\&$ mining in large graphs.
\newblock In {\em Proceedings of the 18th ACM SIGKDD International Conference
  on Knowledge Discovery and Data Mining}, KDD '12, pages 1231--1239, New York,
  NY, USA, 2012. ACM.

\bibitem[\protect\citeauthoryear{Kipf and Welling}{2017}]{Kipf:2016tc}
Thomas~N. Kipf and Max Welling.
\newblock {Semi-Supervised Classification with Graph Convolutional Networks}.
\newblock In {\em Proceedings of the 5th International Conference on Learning
  Representations}, ICLR '17, 2017.

\bibitem[\protect\citeauthoryear{Klimt and Yang}{2004}]{enron}
Bryan Klimt and Yiming Yang.
\newblock The enron corpus: A new dataset for email classification research.
\newblock In {\em Machine Learning: ECML 2004}, pages 217--226, Berlin,
  Heidelberg, 2004. Springer Berlin Heidelberg.

\bibitem[\protect\citeauthoryear{{Li} and {Pan}}{2016}]{lipan}
A.~{Li} and Y.~{Pan}.
\newblock Structural information and dynamical complexity of networks.
\newblock {\em IEEE Transactions on Information Theory}, 62(6):3290--3339, June
  2016.

\bibitem[\protect\citeauthoryear{Minello \bgroup \em et al.\egroup
  }{2018}]{minello}
Giorgia Minello, Luca Rossi, and Andrea Torsello.
\newblock {On the von Neumann entropy of graphs}.
\newblock {\em Journal of Complex Networks}, 11 2018.

\bibitem[\protect\citeauthoryear{Mises and
  Pollaczek-Geiringer}{1929}]{power_iteration}
R.~V. Mises and H.~Pollaczek-Geiringer.
\newblock Praktische verfahren der gleichungsauflösung .
\newblock {\em ZAMM - Journal of Applied Mathematics and Mechanics /
  Zeitschrift für Angewandte Mathematik und Mechanik}, 9(2):152--164, 1929.

\bibitem[\protect\citeauthoryear{Passerini and
  Severini}{2009}]{Passerini2009QuantifyingCI}
Filippo Passerini and Simone Severini.
\newblock Quantifying complexity in networks: The von neumann entropy.
\newblock {\em IJATS}, 1:58--67, 2009.

\bibitem[\protect\citeauthoryear{Perozzi \bgroup \em et al.\egroup
  }{2014}]{perozzi2014deepwalk}
Bryan Perozzi, Rami Al-Rfou, and Steven Skiena.
\newblock Deepwalk: Online learning of social representations.
\newblock In {\em Proceedings of the 20th ACM SIGKDD international conference
  on Knowledge discovery and data mining}, pages 701--710. ACM, 2014.

\bibitem[\protect\citeauthoryear{Ribeiro \bgroup \em et al.\egroup
  }{2017}]{DBLP:journals/corr/FigueiredoRS17}
Leonardo~F.R. Ribeiro, Pedro~H.P. Saverese, and Daniel~R. Figueiredo.
\newblock Struc2vec: Learning node representations from structural identity.
\newblock In {\em Proceedings of the 23rd ACM SIGKDD International Conference
  on Knowledge Discovery and Data Mining}, KDD '17, pages 385--394, New York,
  NY, USA, 2017. ACM.

\bibitem[\protect\citeauthoryear{Scarselli \bgroup \em et al.\egroup
  }{2009}]{Scarselli:2009:GNN:1657477.1657482}
Franco Scarselli, Marco Gori, Ah~Chung Tsoi, Markus Hagenbuchner, and Gabriele
  Monfardini.
\newblock The graph neural network model.
\newblock {\em Trans. Neur. Netw.}, 20(1):61--80, January 2009.

\bibitem[\protect\citeauthoryear{Shervashidze \bgroup \em et al.\egroup
  }{2011}]{Shervashidze:2011:WGK:1953048.2078187}
Nino Shervashidze, Pascal Schweitzer, Erik~Jan van Leeuwen, Kurt Mehlhorn, and
  Karsten~M. Borgwardt.
\newblock Weisfeiler-lehman graph kernels.
\newblock {\em J. Mach. Learn. Res.}, 12:2539--2561, November 2011.

\bibitem[\protect\citeauthoryear{Shetty and Adibi}{2005}]{shetty}
Jitesh Shetty and Jafar Adibi.
\newblock Discovering important nodes through graph entropy the case of enron
  email database.
\newblock In {\em Proceedings of the 3rd International Workshop on Link
  Discovery}, LinkKDD ’05, page 74–81, New York, NY, USA, 2005. Association
  for Computing Machinery.

\bibitem[\protect\citeauthoryear{Tu \bgroup \em et al.\egroup }{2018}]{drne}
Ke~Tu, Peng Cui, Xiao Wang, Philip~S. Yu, and Wenwu Zhu.
\newblock Deep recursive network embedding with regular equivalence.
\newblock In {\em Proceedings of the 24th ACM SIGKDD International Conference
  on Knowledge Discovery \& Data Mining}, KDD ’18. Association for Computing
  Machinery, 2018.

\bibitem[\protect\citeauthoryear{Xinyi and Chen}{2019}]{Xinyi2019CapsuleGN}
Zhang Xinyi and Lihui Chen.
\newblock Capsule graph neural network.
\newblock In {\em ICLR}, 2019.

\bibitem[\protect\citeauthoryear{Xu \bgroup \em et al.\egroup
  }{2018}]{xu2018powerful}
Keyulu Xu, Weihua Hu, Jure Leskovec, and Stefanie Jegelka.
\newblock How powerful are graph neural networks?, 2018.

\bibitem[\protect\citeauthoryear{Ying \bgroup \em et al.\egroup
  }{2018}]{DBLP:journals/corr/abs-1806-08804}
Rex Ying, Jiaxuan You, Christopher Morris, Xiang Ren, William~L. Hamilton, and
  Jure Leskovec.
\newblock Hierarchical graph representation learning with differentiable
  pooling.
\newblock In {\em Proceedings of the 32Nd International Conference on Neural
  Information Processing Systems}, NIPS'18, pages 4805--4815, USA, 2018. Curran
  Associates Inc.

\bibitem[\protect\citeauthoryear{Zaheer \bgroup \em et al.\egroup
  }{2017}]{NIPS2017_6931}
Manzil Zaheer, Satwik Kottur, Siamak Ravanbakhsh, Barnabas Poczos, Russ~R
  Salakhutdinov, and Alexander~J Smola.
\newblock Deep sets.
\newblock In {\em Advances in Neural Information Processing Systems 30}, pages
  3391--3401. Curran Associates, Inc., 2017.

\bibitem[\protect\citeauthoryear{Zhang \bgroup \em et al.\egroup
  }{2018}]{Zhang2018AnED}
Muhan Zhang, Zhicheng Cui, Marion Neumann, and Yixin Chen.
\newblock An end-to-end deep learning architecture for graph classification.
\newblock In {\em AAAI}, 2018.

\end{thebibliography}
\bibliographystyle{named}

\end{document}